# A Novel IoT-based Framework for Non-Invasive Human Hygiene Monitoring using Machine Learning Techniques

**Md Jobair Hossain Faruk***, **Shashank Trivedi**†, **Mohammad Masum**†, **Maria Valero**†‡
**Hossain Shahriar**†, **Sheikh Iqbal Ahamed**§

*Department of Software Engineering and Game Development, Kennesaw State University, USA
†Department of Information Technology, Kennesaw State University, USA
§Department of Computer Science, Marquette University, USA

‡Corresponding Author: mvalero2@kennesaw.edu

***Abstract*—Peoples' personal hygiene habits speak volumes about the condition of taking care of their bodies and health in daily lifestyle. Maintaining good hygiene practices not only reduces the chances of contracting a disease but could also reduce the risk of spreading illness within the community. Given the current pandemic, daily habits such as washing hands or taking regular showers have taken primary importance among people, especially for the elderly population living alone at home or in an assisted living facility. This paper presents a novel and non-invasive framework for monitoring human hygiene using vibration sensors where we adopt Machine Learning techniques. The approach is based on a combination of a geophone sensor, a digitizer, and a cost-efficient computer board in a practical enclosure. Monitoring daily hygiene routine may help healthcare professionals be proactive rather than reactive in identifying and controlling the spread of potential outbreaks within the community. The experimental result indicates that applying a Support Vector Machine (SVM) for binary classification exhibits a promising accuracy of ~95% in classification of different hygiene habits. Furthermore, both tree-based classifier (Random Forrest and Decision Tree) outperforms other models by achieving the highest accuracy (100%), which means that classifying hygiene events using vibration and non-invasive sensors is possible for monitoring hygiene activity.**

## I. Introduction

The world is aging rapidly and the number of elderly people would be outnumbered compared to children. According to the World Health Organization (WHO), the number of people over the age of 60 years surprisingly will be doubled nearly from 12% to 22% in the next 30 years due to the increasing life expectancy gradually for the medical and technological advancements over the last few decades [1]. According to the US Census Bureau reported in 2016 [2], [3], the US population is aging slower than the rest of the world where the number of Americans ages 65 and older is projected to nearly double from 52 million in 2018 to 95 million by 2060. Although it is great news that people are living longer, however, a longer life does not necessarily mean a healthier life. Aging is very relative and indirectly influenced by several socioeconomic factors such as education, wealth, and employment. Typically, developed nations such as Australia, Canada, and the US often have longer life expectancy than other developing nations due to better economies and better healthcare systems. The US spends more on healthcare, nearly twice as much as the average country that belongs to the Organisation for Economic Co-operation and Development (OECD) [4].

Throughout the aging process, a variety of molecular and cellular damages occur, and they cause not only physical and mental deficiency but also a growing risk of disease, and ultimately, death [5]. Such phenomenon can neither be stopped, delayed, or avoided; it is a natural progression of a human's life-cycle in which the only option is acceptance. The research community has constantly been working to improve the age's challenges by identifying the major concerns followed by finding possible solutions towards increasing life expectations. Improving both physical and mental states is only the initiation where any factors including housing, treatment, relocation, and caregiving needs to be taken into consideration towards finding better technologies for making the aging process easier.

The steady rise in the elderly population results in an increased demand for nursing homes and assisted living facilities. Unfortunately, the elderly population tends to have the highest disability rate, which maximizes the need for long-term care services and daily activities assistance [6]. Long-term care is defined as a variety of services designed to meet a person's health or personal care needs during a short or long period of time and designed to help elderly people in particular within an assisted living facility such as nursing homes or adult daycare homes [7]. In 2016, CDC [8] reported 15,600 nursing homes and 1.7M licensed beds in the United States. Even though these nursing and living assistance homes provide services by qualified personnel, it is always a challenge to monitor the seniors in privacy contexts; for example, restroom activities. Although the senior lives at home with family members, the time to go to the bathroom is always private, making the process of monitoring in this scenario is always uncomfortable for the seniors. Thus, finding a technological solution within non-invasive concepts for keeping the care of the senior population in uncomfortable scenarios needs researchers' attention.

Leveraging the advanced technology to address real-world scenarios and equipping the healthcare workers with the right set of tools is crucial in order to pave the healthcare system to be a proactive force rather than a reactive one. People's daily habits indicate a lot about their health and well-being. According to the Centers for Disease Control and Prevention (CDC) [9], researchers in London estimate that if everyone routinely washed their hands, a million deaths a year could be prevented. Questions such as "are we washing your hands regularly?" or "are we taking enough showers?" may not seem as important during the early adult years, but these questions become essential with aging as it could be an indication of developing health crises such as dementia. The elderly community is at the most risk where the majority of them either live alone at home or live in an assisted living facility where the staff may be stretched too thin. To avoid another potential outbreak of a deadly disease, ensuring proper hygiene should be the primary concern, and evaluating the existing monitoring techniques is a demand towards improving the potential methods that can be more effective in improving the health, lifestyle, and quality of life.

In this study, we propose to leverage the power of not only sensors technology but also machine learning to help better understand the daily habits of the elderly population. Machine learning is a widely used method that enables computers to imitate and adapt human-like behavior which has been applied to various mundane and complex problems [10]. Machine learning algorithms including Decision Tree (DT), Random Forest (RF), Naive Bayes (NB), Logistic Regression (LR) as well as Neural Network (NN)-based classifiers have been adopted in many fields including smart children management, economics and finance, brain-computer interfaces (BCI), malware and ransomware detection [11]–[15]. Adopting machine learning in this experiment shall enable to measure and improve the accuracy of the Hygiene Monitoring system which will help healthcare workers, and family members, to take proactive measures in remediating a potential health crisis. The primary contributions of the paper are as follows:

- We present a novel and non-invasive framework for monitoring human hygiene using non-intrusive vibration sensors and machine learning.
- We demonstrate the proposed framework by conducting robust experiments and comparing it with the various ML classification methods.
- We show the possibility to extract accurate information about the type of hygiene task performed by a senior in a private setting, the bathroom for instance, without using cameras or other invasive methods

The rest of this paper will be organized as follows: In Section II, we discuss the related work about human hygiene monitoring using vibration sensor technology. Section III presents the proposed framework and the system architecture that enables monitoring human hygiene activities while Section IV describes the experimental setting and results of the proposed approach. Section V discusses the challenges, limitations, and future work. Section VI concludes the paper.

## II. Literature Review

We adopted a systematic analysis to identify previous studies in the area of non-invasive monitoring using vibration sensors and classifiers [10]. Based on the study, we identified that none of the existing approaches used the combination of the geophone, digitizer, and computer board (raspberry pi) to measure multiple activities in the bathroom (sink use and toilet use) or kitchen sink used for hand wash. In this section, we discuss several papers and works that are closely related and served as an inspiration for this research work.

Yiyuan Zhang *et al.* [16] proposed a wearable accelerometer to monitor six types of bathroom activities: dressing, undressing, washing hands, washing face, brushing teeth, and toilet. Two main models for validation were used: 10-fold cross-validation and LOPO (Leave-One-Person-Out) model where all participant data was shuffled and split into 10 folds where one-fold was selected for testing while the other 9 folds were used for training. In the LOPO model, the dataset for a participant was used for testing while the other participant's data were used for training. The outcome of this research concluded that a wearable device could be used to monitor six-bathroom activities but further studies are suggested due to the lower classification score on some activities including washing face and undressing. Additional research using this technique is recommended to improve the success ratio for the activities in the lower classification category. *Our proposed framework is different as the person is not required to use any device, which is especially important for the elderly population.*

Jianfeg Chen *et al.* [17] focuses on the strength of acoustic signals to monitor Activities of Daily Living (ADL) in the elderly that are designed to recognize and classify different activities occurring within a bathroom based on the sound. The system architecture includes an infrared system that would detect entry into the bathroom, a microphone with a pre-amplifier circuit and a laptop is required to record and classify the sound events in real-time using Hidden Markov Models (HMMs). A personalized daily report is generated by the system which provides a daily summary of the subject's bathroom behavior and provides recommendations based on the data recorded. The study resulted in 87% or higher accuracy in detecting activities and event classification that shall improve the accuracy of the event recognition in the future to enhance the system's capability to monitor and identify human vocals to assess the subject's state of mind within the bathroom. *Our proposed work is different as we use vibration signals as a primary and unique form of information. We also process the data inside a Raspberry Pi, which is about the location of laptops inside bathrooms.*

Yingqi Hao *et al.* [18] created a safety monitor that not only detects a safety hazard in the kitchen but will also respond to the threat and send out alerts or notifications. The system utilizes a single-chip microcomputer as the central core along with sensors to detect temperature, gas, and smoke concentration. On the software end, it uses the C-language to control the MCU along with Keil C51 for debugging purposes. It creates a threshold for normalcy and any abnormal behavior that crosses the threshold will trigger an action from the system. The proposed infrastructure combined with the hardware used for research in this paper can create a robust "monitoring and response" system in case of any potential hazard performing ADLs. *The system is promising and was an inspiration for our study, we analyzed different phases where we used various types of data (vibration) as the source.*

K. Jimi *et al.* [19] proposed an approach using a 79GHz millimeter-wave UWB sensor with Fast-Chirp Modulation (FCM) and hidden Markov model (HMM) for bathroom monitoring. A non-intrusive monitoring method is adopted as the bathroom raises the concerns of privacy infringement, humidity, and fluctuations in the water levels. By installing the radar sensor on the wall, it will allow for the velocity, angle, and distance to be measured. The experiment was run in three separate scenarios which included a normal bathroom without any accident in scenario A, drowning in the bathtub for scenario B, and falling in the washing area for scenario C. The results showed a 95% success rate for the prior learning method and a 90% success rate for the incremental learning approach. *In our study, we use a geophone that senses "z" directions and with a low sampling rate (100Hz).*

Jonathon Fagert *et al.* [20] focused on the four activities associated with the event of hand-washing including steps, water, soap, and rinsing. The approach also utilizes recording and measuring the structural vibration signals through the means of a geophone. The vibration signal is then converted to digital using ADC (Analog-to-Digital) converter and processed using a computer. The classification of the activities is conducted through the SVM and utilizes the L-1 norm for efficiency. The technique used for classification is called "one-against-one" in which six SVM models are used to label each activity considered as approved or classified when 3 of the 6 models classify as the same label. Overall, the result shows positive with a 95.4% average accuracy which is a 1.4X improvement over the baseline of 69.1 percent. *This study is the most closely related approach to our proposed work, as structural vibration is obtained and analyzed. However, besides the hand-washing process, we also analyze the use of the toilet and the kitchen sink. Furthermore, we provide analysis with different machine learning approaches to identify the one(s) with the highest accuracy.*

Jeroen Klein Brinke *et al.* [21] employs the use of radio-frequency-based recognition to monitor the variations in several shower-related activities. The experiment was demonstrated in a regular/normal shower and an instance of the troublesome shower for comparison. The experimental setup included multiple custom Gigabyte Brix IoT devices with Intel Apollo Lake processor, 8GB RAM, and Intel N Ultimate network card where one acted as transmitter and the other as a receiver. A Convolutional Neural Network (CNN) was used for the classification of the dataset as it preserves both the structural and spatial information. Although the experiment was a success in determining whether the shower is on or off, the authors do recommend validating the results in the paper with further trials which would include more participants, different locations, and different frequencies in new settings.

## III. System Architecture

We present a four-stage approach to the system architecture including (i) Data Collection, (ii) Data Validation, (iii) Data Extraction, and (iv) Data Analysis shown in Fig. 1.

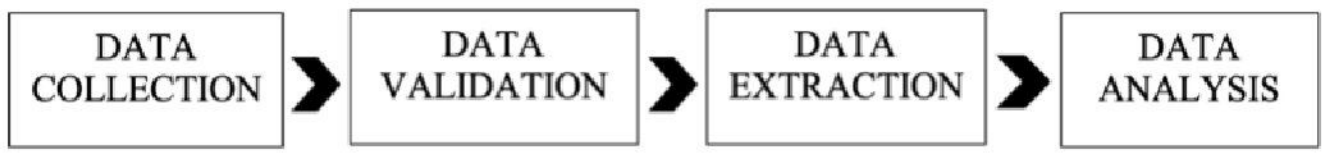

Fig. 1. Four Stages of system architecture

### A. Data Collection

In this first stage of the experiment, we focused on identifying the type of sensor required to collect the data and the type of data to be collected including the location of the sensor placement and the data points that are most relevant to the research. We started with a Raspberry Shake turnkey product display in Fig. 2 which included a Raspberry circuit board, a smart sensor composed by a geophone (vertical axis seismic sensor), a digitizer board, and a simple-board computer (Raspberry Pi 3B) all in a preassembled enclosure [22].

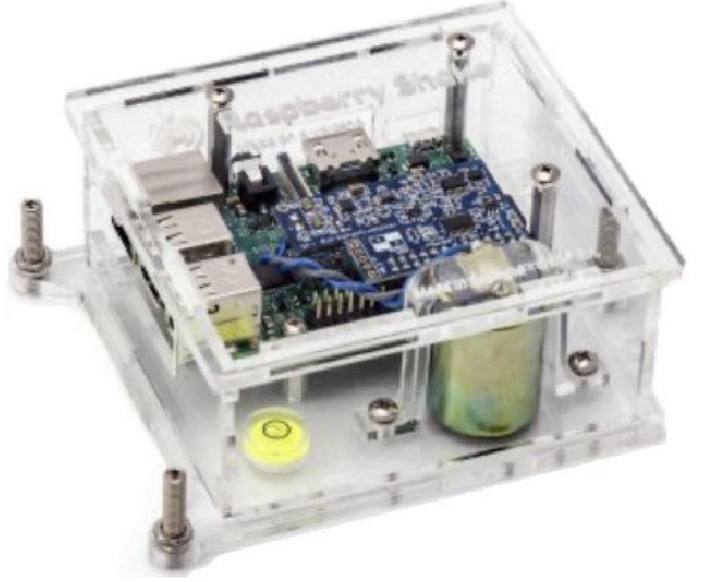

| Raspberry Shake I/O |
|---|
| 1 x Ethernet Port |
| 1 x HDMI Port |
| 4 x USB Ports |
| 1 x 3.5MM jack |
| 1 x MicroSD card slot |

Fig. 2. Raspberry Shake Enclosure and Raspberry Shake I/O Ports

The geophone (Figure 4) are highly sensitive motion transducers that have been used by seismologists and geophysicists for decades. In its simplest form, it is just a coil suspended around a permanent magnet as like a loudspeaker coil/magnet system. When the coil moves relative to the magnet, a voltage is induced in the coil according to the Faraday law [24]. The induced voltage is proportional to the relative speed, thus the geophone sensor element is measuring velocity, which we call here vibration.

Before getting started with data collection, the Raspberry Pi was connected to a network while a dashboard was configured in Grafana for data visualization [25]. To get the Raspberry Pi configured with network settings, we connect it to an external monitor using an HDMI cable and also connect a wired keyboard to run the commands. At the Command-Line, we log in with credentials and run the following command to access the network interface to add the own network details (SSID and passphrase).

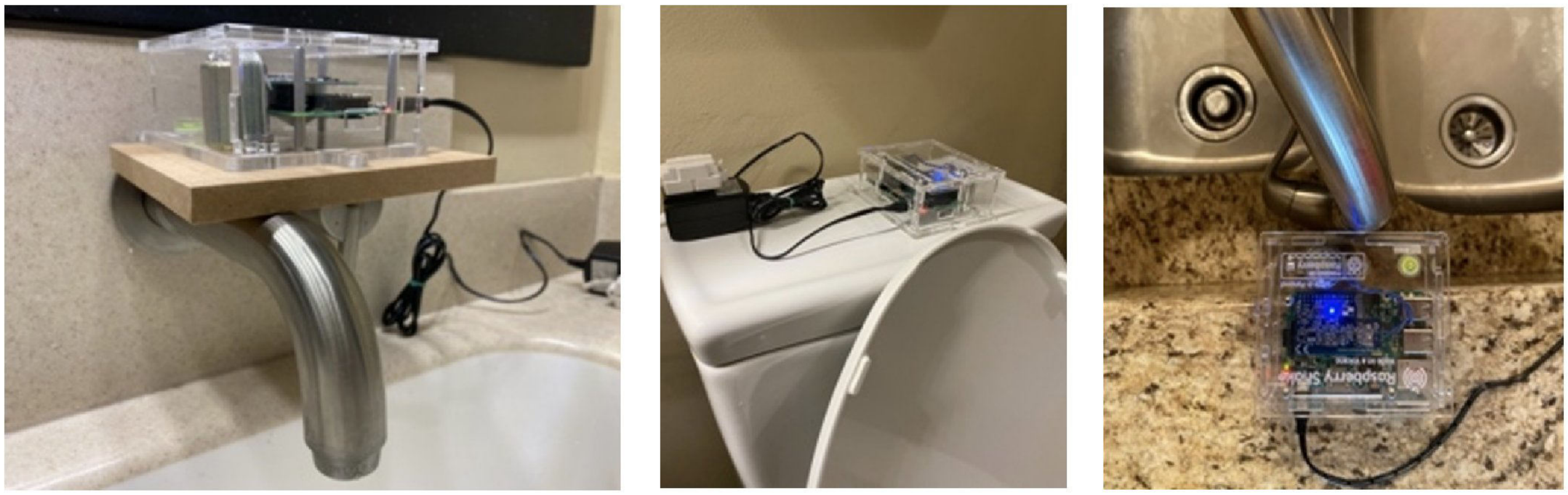

Fig. 3. Bathroom Faucet (left), Toilet (middle), and Kitchen Sink (right) Monitoring Setup

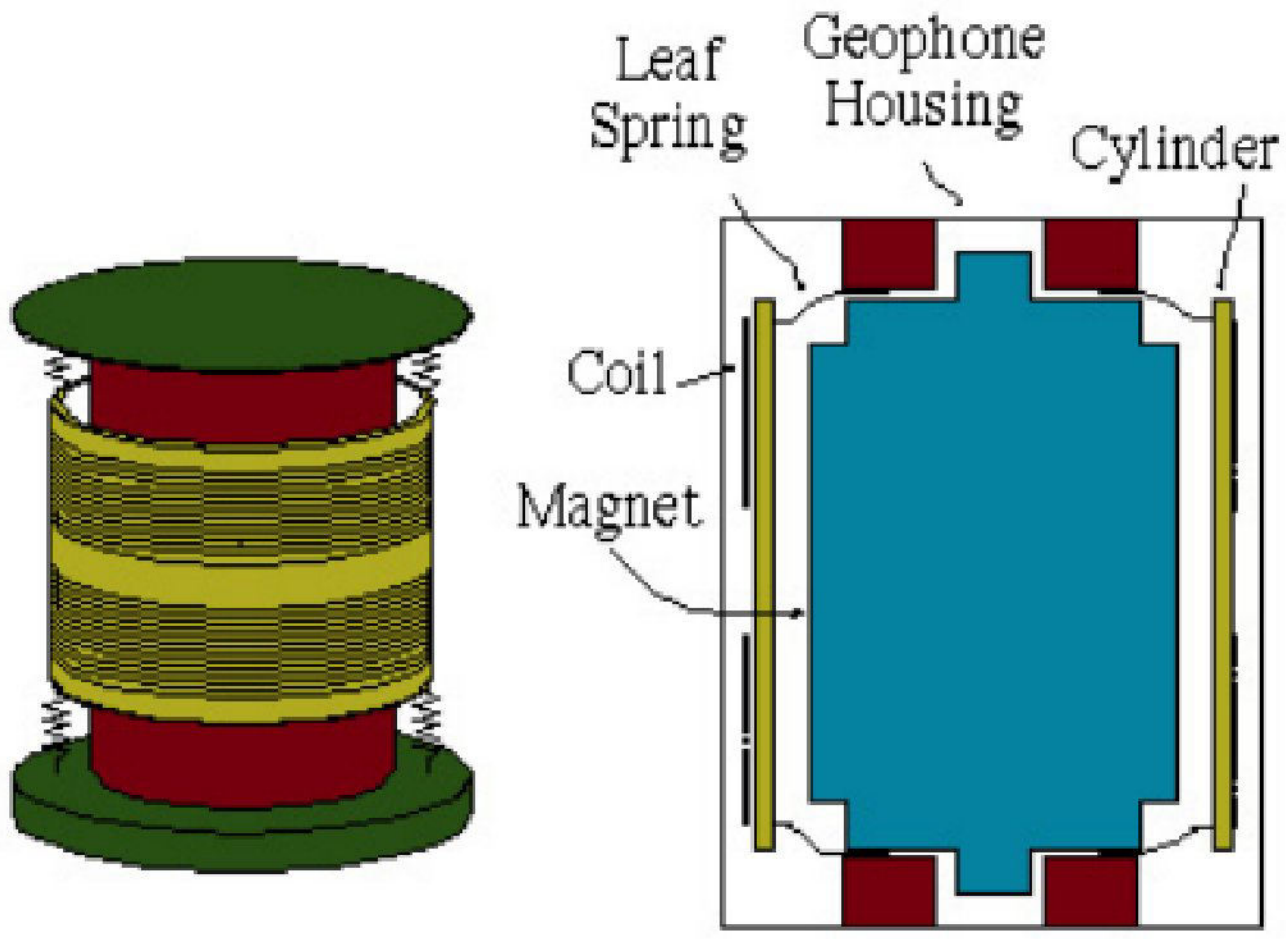

Fig. 4. An isometric and cross-sectional view of a typical geophone, after Aaron Barzilai [23]

After completing the configuration, the device starts recognizing the network and joins automatically to configure a local instance of Grafana for visualizing the data. A unique IP address of the Raspberry Pi database port was used. After configuring the data source, we connect InfluxDB (time-series database) [26] and set up the Raspberry Pi, Grafana, and InfluxDB together. We identify various locations for the sensor placement and also identify some key data points that are relevant and can add value to this research. In accordance with the primary goal to monitor human hygiene using vibrations from water, we chose to narrow down the scope and focus on two of the six ADLs including (i) personal hygiene (washing hands) and (ii) toileting (flushing toilet) [27]. Monitoring the vibrations through the use of the toilet was the first location under the ADL category and the other two locations were the bathroom faucet and kitchen sink of personal hygiene. Figure 3 shows the location of the sensor on the bathroom and kitchen sink and bathroom faucet.

For the scope of this study, we assume that the subject is washing their hands each time the bathroom faucet or the kitchen sink is opened and closed. All the associated activities are related to the concept of human hygiene and include capturing vibrations from water. While considering the sensor placement, it was also important to consider the environment the sensor is in which may include the possibility of it getting wet or challenges that come with increased humidity where we also consider factors related to external interference such as people walk, sensor falling off due to off-balance, or sensor is getting pulled accidentally. Due to the size and shape of the faucet, we made a makeshift solution to hold the sensor on top of the bathroom faucet as shown in Figure 3 (left). For the kitchen sink, the sensor was placed on the side of the faucet, as shown in Figure 3 (right). Figure 3 (middle) shows the sensor placed on top of the toilet tank.

After identifying the locations of sensor placement, we figure out the key data points that should be collected which can help understand how the sensor was placed and used. These data points needed to be logged for the purpose of data validation (as part of stage two)for determining whether or not vibration sensors and classifiers can be used to track the usage of ADLs. The validation phase plays a critical role in verifying the data stored on the device with the data logged manually through the life of the experiment. It was also important for the data logging to be as accurate as possible so that it aligns as closely with the actual as possible in determining the success of this study.

We record various data points for each location as part of the experiment log includes: (i) Date: Date of the recorded event, (ii) Start Time: When the event began (iii) Duration of the event: How long the event lasted for (iv) Building Type: Type of building where the sensor was placed, (v) Location of the sensor: Where was the sensor placed within the building,

| Date | Start Time | Duration (in sec) | Bldg. Type | Location | Position | Event Type |
|---|---|---|---|---|---|---|
| 6/14/2021 | 08.24 AM | 50 | Residential | Bathroom Faucet | Top | Opening/Closing Faucet |
| 6/14/2021 | 08.28 AM | 15 | Residential | Bathroom Faucet | Top | Opening/Closing Faucet |
| 6/14/2021 | 08.31 AM | 20 | Residential | Bathroom Faucet | Top | Opening/Closing Faucet |
| 6/14/2021 | 08.35 AM | 65 | Residential | Bathroom Faucet | Top | Opening/Closing Faucet |
| 6/14/2021 | 09.08 AM | 12 | Residential | Bathroom Faucet | Top | Opening/Closing Faucet |
| 6/14/2021 | 09.15 AM | 18 | Residential | Bathroom Faucet | Top | Opening/Closing Faucet |
| 6/14/2021 | 09.33 AM | 10 | Residential | Bathroom Faucet | Top | Opening/Closing Faucet |
| 6/14/2021 | 11.00 AM | 25 | Residential | Bathroom Faucet | Top | Opening/Closing Faucet |
| 6/14/2021 | 12.00 PM | 20 | Residential | Bathroom Faucet | Top | Opening/Closing Faucet |
| 6/14/2021 | 12.44 PM | 2 | Residential | Bathroom Faucet | Top | Opening/Closing Faucet |
| 6/14/2021 | 02.44 PM | 60 | Residential | Bathroom Faucet | Top | Opening/Closing Faucet |
| 6/14/2021 | 05.54 PM | 30 | Residential | Bathroom Faucet | Top | Opening/Closing Faucet |
| 6/14/2021 | 06.28 PM | 10 | Residential | Bathroom Faucet | Top | Opening/Closing Faucet |
| 6/14/2021 | 07.25 PM | 20 | Residential | Bathroom Faucet | Top | Opening/Closing Faucet |

Fig. 5. Sample of Data Log from Bathroom Faucet

(vi) Position of the sensor: How was the sensor placed at its location, (vii) Event Type: How did the subject interact with the sensor, and (viii) Sensor Distance (approx.) from the water source: Distance from the water source.

A total of 368 events were captured from all three locations over several weeks of the experiment to conduct analysis and draw any applicable conclusions. This data was logged manually on a paper and then documented electronically in a spreadsheet as shown in Figure 5. During the data validation stage, the actual data of the sensor will be verified against this data logged to measure the accuracy of recorded data and provide validity to the experiment where a comparison between what the device recorded versus what actually happened. Throughout the data collection stage, we had to overcome various challenges including the shape and size of the device and faucet, open ports (water intrusion), failure to log the time, and adjust the intensity of the water flowing through the faucet.

We adopt the qualitative research methodology to get the right data than to collect abundant data. The accuracy of the collected data, in this case, gives this research more meaning than the quantity of the data; this is especially important during the analysis stage as the quality of the data is far more critical in training the machine learning models than the quantity. Data was collected in a controlled environment over a period of about 33 days between June 14 and August 20, 2021. The prototype device was placed by the bathroom sink, toilet tank, and kitchen sink to simulate a real-world need. All the data was collected using the same single Raspberry Shake moved between the three locations [22].

### B. Data Validation

Data Validation is another important stage of the process as it allows for the data to be validated through the means of comparison between what the device captured vs what actually happened. In this stage, there are two main components including viewing the data captured on the device and comparing that data with the manual log that was maintained. This data validation stage is actually in effect through the life of the experiment where every time the location of the device is altered or if the device is rebooted for any reason, we will need to verify if the data is still being recorded.

### C. Data Extraction

After collecting enough data, we extract the data for analysis purposes where extracting the data is a simple yet tedious process as it requires further validation on each individual event. This data extraction is completed by accessing Grafana on the device itself where data is exported and saved at a local computer location. We then apply a band-pass filter [28] to eliminate the surrounding noise as much as possible that will improve the quality of data.

### D. Data Analysis

All the extracted data is imported into MatLab in CSV formats exported to read the data first and then extract some features from that data that can potentially help us train the machine learning algorithm in detecting the events and classifying them as one of the ADLs. It is important to note that not every feature will be applicable in the classification of the event. A number of trials and errors shall be found in the process where figuring out which feature is able to be used for classification purposes is crucial. After importing the data, we adopt mapping techniques for verification and confirmation purposes followed by plotting the data in MatLab, the corresponding graph should be an exact replica of Grafana, Figure 7 shows a comparison between the data plotted in Grafana (top) and MatLab (bottom).

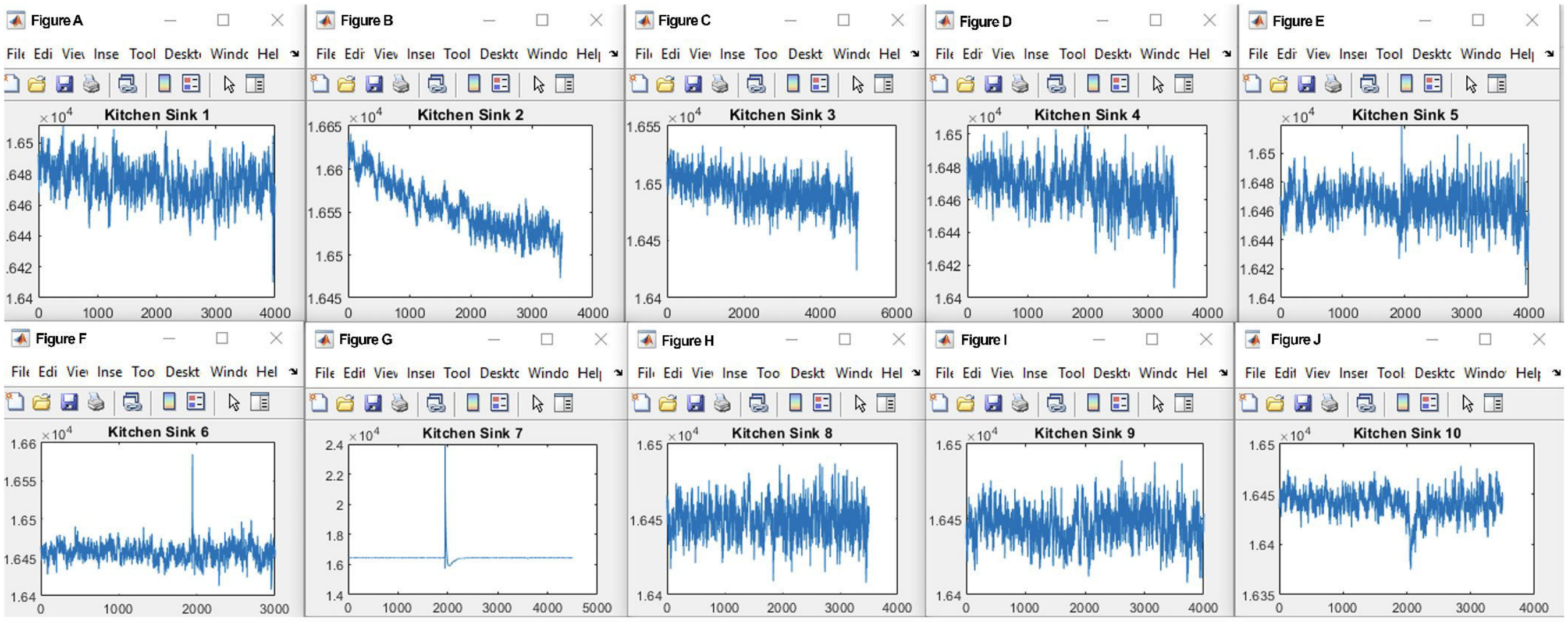

Fig. 6. Comparing Samples of the Same Event

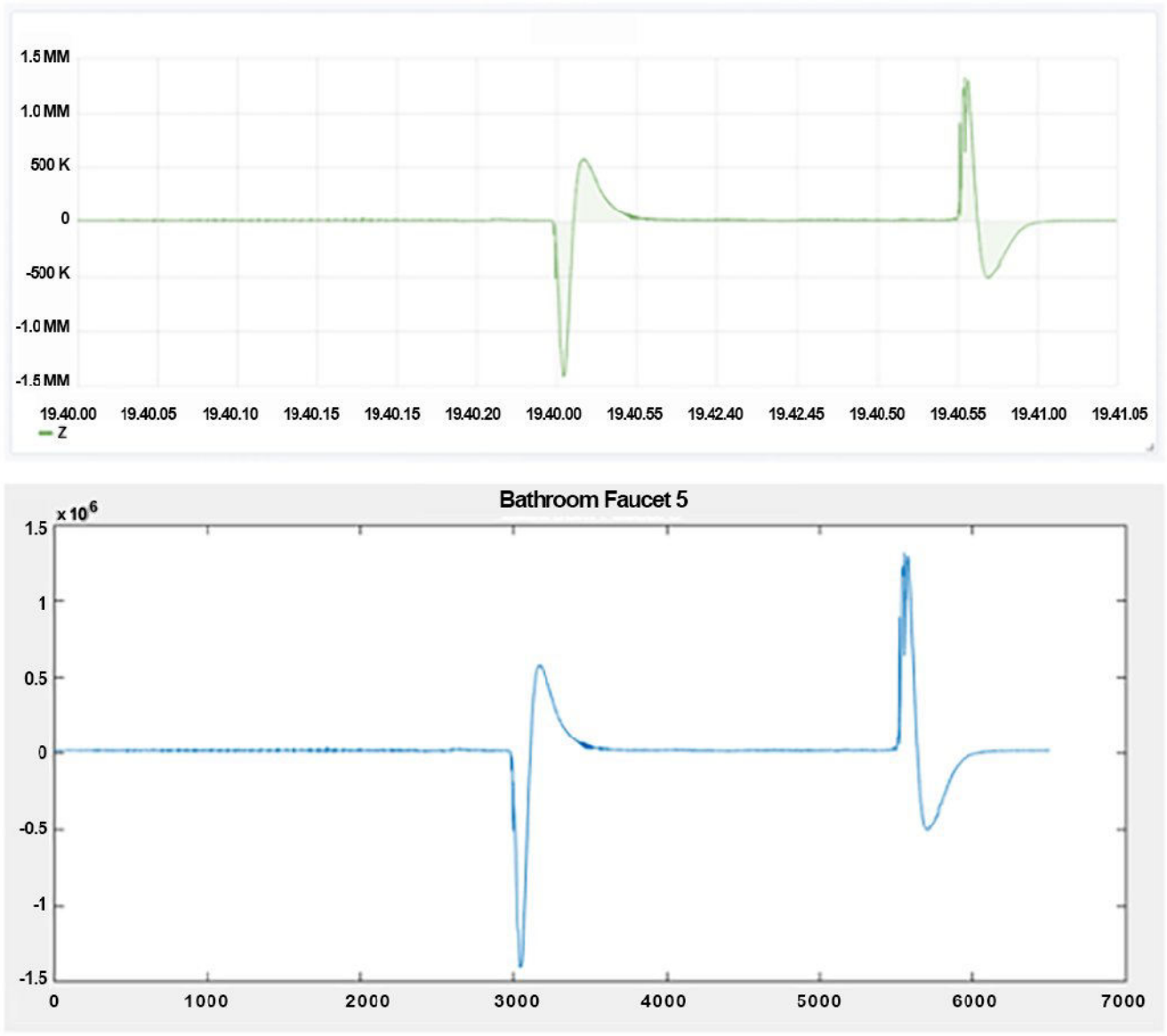

Fig. 7. Comparing Data Plotting in Grafana (top) and MatLab (bottom)

After plotting and verifying the data, we applied feature extraction techniques that allow us to manipulate the data in a way that can highlight the similarities and differences between the datasets. Feature extraction is a critical step not only to interpret and comprehend the data, but it is also necessary for the Support Vector Machine (SVM) as the results from this feature extraction will be fed into the machine learning algorithm for the automated classification of the hygiene event. Without feature extraction, it will be difficult to process the raw data as that is only values tied to a point in time. These features need to be relevant to the data and activity. The list of features that were extracted is illustrated in Table I.

TABLE I
LIST OF EXTRACTED FEATURES

| | |
|---|---|
| Kurtosis | Location of the Highest Peak |
| Standard Deviation | Location of the Lowest Peak |
| Entropy | Peak Difference |
| Highest Peak | Mean |
| Lowest Peak | Period |

## IV. EXPERIMENT & RESEARCH FINDINGS

During the analysis phase, we realized that the quality of data is as important as the quantity of data; for instance, we can have 1000 events captured, but if these 1000 events do not correspond accurately to the log that was maintained then the possibility of extracting incorrect data shall be increased and which would then lead to incorrect feature identification resulting in inconsistent findings. All the issues encountered during the initial capture resulted in a lack of confidence with the first batch of data, therefore, additional data was collected for verifying each event in real-time to increase the accuracy and credibility of the data. The new batch of data included 30 new events which were further broken down into a new category of "Behavior" that would indicate the level or amount of faucet opened which represents the "intensity" of the water through the faucet. The very first step on the data is to plot out the events to visualize the similarities and differences between the various samples for the same event at the same location. This visualization, as shown in Figure 6, of the data would help in better understanding how factors such as angle, intensity, and even duration could influence the dataset. It can also help find anomalies such as Kitchen Sink seven events in Fig. 6.

As we compiled the results from feature extractions on each of the hygiene events, it was pretty apparent to a human eye which features would be ideal for autonomous identification and classification, but the SVM would not be able to differentiate at a glance as like how a human can. In order to gain such capability to distinguish between the hygiene events by the SVM, each of the values from the feature extraction would need to be fed into the machine learning model for training and cross-validation. Some of the initial observations from the analysis of the features as follows:

- Most of the values seem to have fallen within a range, except for a few outliers: this is to be expected as the results would need to be somewhat consistent for event classification.
- The standard deviation for the bathroom faucet was very high: this was also expected as the sensor was sitting on top of the faucet and the opening/closing of the faucet would have created increased movements.
- The highest and lowest peak numbers were also distinct for the bathroom faucet as it involved the movement of the sensor with each opening/closing event which would have resulted in high/low frequencies.

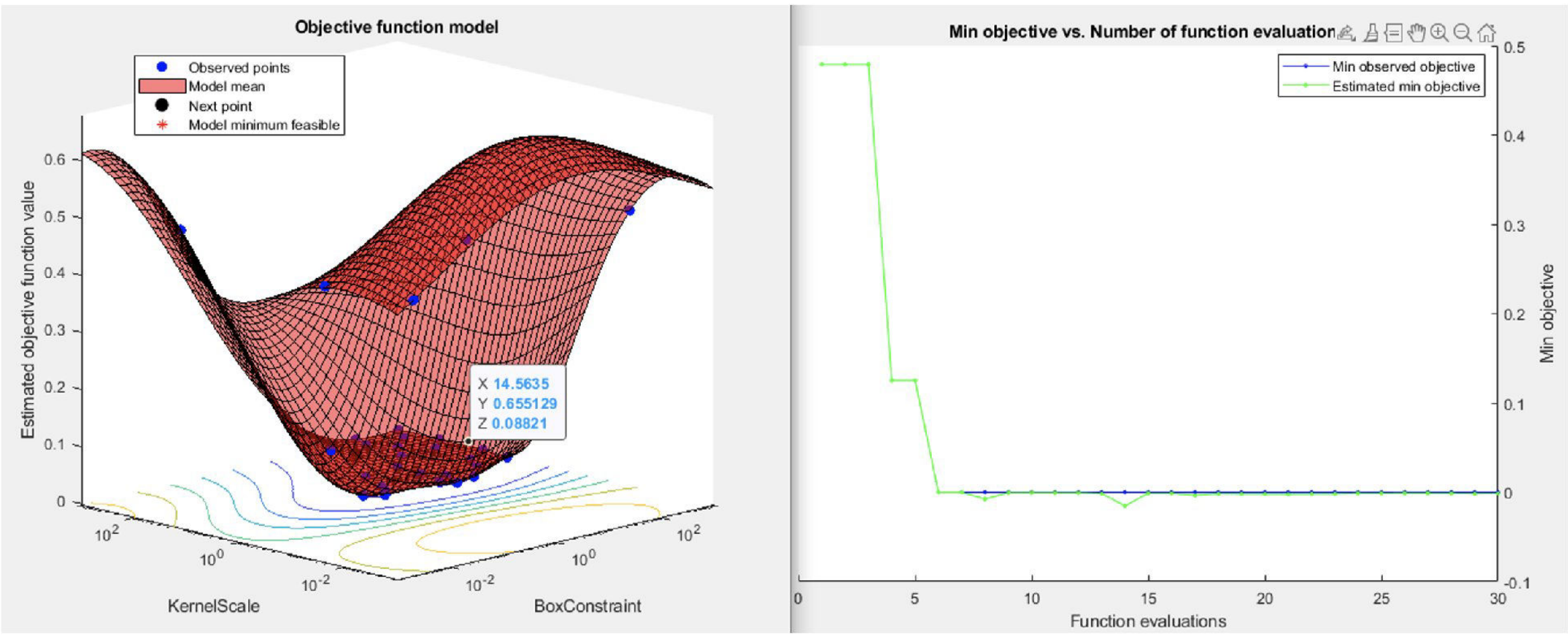

Fig. 8. Objective Function Model

Support vector machine (SVM) is a machine learning-based supervised method for solving classification and regression problems using a hyperplane in N-dimensional space towards effective nonlinearity and high dimension solution [29]. This is why feature extraction is a necessary step in the classification via the SVM as it emphasizes the data points that add the most value and could be used in the determination of the support vectors. In this study, we will be using a binary classification schema which is an OR model that will help us differentiate between two classes which are the different hygiene events and/or locations. Before diving into the specifics of the configuration of the classification model, the dataset will need to be reformatted in a way that is readable by the learning model. The data will need to be split in two ways: (i) Labels: numeric values corresponding to the three locations including kitchen, bathroom, and toilet of the events and (ii) Values: the results of the feature extractions from each of the events. Both files need to be saved in Comma Separated Values (CSV) format for importing into MatLab.

When creating these files, consistency is the utmost importance as the values in the second file need to be in exact order to accurately correspond to the labels in the first file. The features shall be lined up accordingly so that SVM does not read jumbled data to avoid the throwing of the classification model when the values are mixed. Since we used a binary classification model, we processed three separate files containing 60 events each which shall compare data from the kitchen sink with bathroom faucet, bathroom faucet with toilet flushing, and toilet flushing with kitchen sink to measure the accuracy of the classification from each batch.

Next would be to divide up the dataset into training and test batches where the training set would be the larger set of the two as we need more data to train the model than to test. We adopt an 80:20 ratio which divided 24 events from each location for training and 6 events were left for testing the trained set where each file consists of 48 events for training and 12 for the test. To further increase the accuracy of results, the data was split in a random fashion which means the algorithm was responsible for choosing which data to test at random. The function of the random number generator was used which divided up the events in a random order for both the training and test set. Furthermore, the cross-validation set was prepared with a K-5 fold to train using the entire dataset to improve accuracy by minimizing errors followed by processing the data where it allows the SVM to choose the three best features out of the ten features used for input, more features do not necessarily mean more accuracy. Features extracted that are relevant to the kind of data that are working with, but the features themselves do not add much value to the classification. For each of the three datasets used for comparison, the SVM chose separate features as prominent for classification. No predefined features were selected and the

**Kitchen Sink and Bathroom Faucet - 95%**

| Attempt | Features | Accuracy | Input | Misclassification |
|---|---|---|---|---|
| 1 | 1, 2, 10 | 100% | 4KS, 8BF | 0 |
| 2 | 1, 2, 10 | 83.33% | 8KS, 4BF | 2 |
| 3 | 1, 2, 10 | 100% | 6KS, 6BF | 0 |
| 4 | 1, 2, 10 | 91.67% | 4KS, 8BF | 1 |
| 5 | 1, 2, 10 | 100% | 8KS, 4BF | 0 |
| 6 | 1, 2, 10 | 83.33% | 6KS, 6BF | 2 |
| 7 | 1, 2, 10 | 100% | 6KS, 6BF | 0 |
| 8 | 1, 2, 10 | 91.67% | 6KS, 6BF | 1 |
| 9 | 1, 2, 10 | 100% | 7KS, 5BF | 0 |
| 10 | 1, 2, 10 | 100% | 4KS, 8BF | 0 |

**Bathroom Faucet and Toilet Flushing - 93.33%**

| Attempt | Features | Accuracy | Input | Misclassification |
|---|---|---|---|---|
| 1 | 1, 9, 10 | 91.67% | 6BF, 6TF | 1 |
| 2 | 1, 9, 10 | 91.67% | 7BF, 5TF | 1 |
| 3 | 1, 9, 10 | 83.33% | 4BF, 8TF | 2 |
| 4 | 1, 9, 10 | 100% | 6BF, 6TF | 0 |
| 5 | 2, 9, 10 | 100% | 8BF, 4TF | 0 |
| 6 | 1, 9, 10 | 100% | 7BF, 5TF | 0 |
| 7 | 1, 9, 10 | 100% | 6BF, 6TF | 0 |
| 8 | 1, 9, 10 | 83.33% | 5BF, 7TF | 2 |
| 9 | 1, 9, 10 | 100% | 8BF, 4TF | 0 |
| 10 | 1, 9, 10 | 83.33% | 6BF, 6TF | 2 |

**Kitchen Sink and Toilet Flushing - 99.17%**

| Attempt | Features | Accuracy | Input | Misclassification |
|---|---|---|---|---|
| 1 | 1, 2, 10 | 100% | 8KS, 4TF | 0 |
| 2 | 1, 2, 10 | 100% | 4KS, 8TF | 0 |
| 3 | 1, 2, 10 | 100% | 6KS, 6TF | 0 |
| 4 | 1, 2, 10 | 100% | 3KS, 9TF | 0 |
| 5 | 1, 2, 10 | 100% | 8KS, 4TF | 0 |
| 6 | 1, 2, 10 | 100% | 8KS, 4TF | 0 |
| 7 | 1, 2, 10 | 100% | 6KS, 6TF | 0 |
| 8 | 2, 9, 10 | 100% | 7KS, 5TF | 0 |
| 9 | 1, 2, 10 | 91.67% | 6KS, 6TF | 1 |
| 10 | 1, 2, 10 | 100% | 6KS, 6TF | 0 |

Fig. 9. Comparing the Success Ratio from 10-Run Trial

SVM was allowed to choose the features that it selected as the most ideal for classification.

TABLE II
GENERALIZED TABLE FOR SEARCH CRITERIA

| LOCATION | SVM SELECTED FEATURES |
|---|---|
| Kitchen Sink: Bathroom Faucet | [1, 2, 10] Kurtosis, Standard Deviation, Period |
| Bathroom Faucet: Toilet Flushing | [1, 9, 10] Kurtosis, Mean, Period |
| Toilet Flushing: Kitchen Sink | [1, 2, 10] Kurtosis, Standard Deviation, Period |

The algorithm finds the features that are the most ideal to create the hyperplane which is essentially the decision boundary with maximum margins possible to predict successfully and classify with the highest theoretical accuracy. Since the approach is binary classification, we used "fitcsvm" to build the model with the best hyperparameters. This allows us to use kernel functions such as Radial Basis Function (RBF) kernel as the learning mechanism, calculated as Equation 1.

$$G(X_j, X_k) = exp(-||X_j - X_k||^2 \tag{1}$$

Support Vector Machines utilize kernel functions to find the support classifiers within a higher dimension. In this approach, the RBF relies on observations that are closest to the ones that we attempt to classify for prediction purposes. The goal here is to minimize the loss during cross-validation of the data. Fig. 8 illustrates how the model evaluates 30 iterations to try and get as close to no errors as possible.

The model was cycled 10 times, as a result, a total of 120 events for each of the three classification scenarios are used. Overall, the training model allows us a very consistent and high accuracy each time with an overall accuracy of above 90% as depicted in Fig. 9. These results help demonstrate the true potential and the value this approach can add to the end goal of automating the detection of hygiene events to make the healthcare professionals be proactive.

The proposed framework distinguishes the various activities based on vibrations only. Fig. 10 shows the success ratio through the confusion matrix for each of the three comparison sets. Between kitchen and bathroom, six of the kitchen events were misclassified as bathroom giving us a 95% success ratio for classification. For bathroom faucet and toilet flushing, four of the faucet events were miscategorized as the toilet, while 4 of the toilet events were miscategorized as faucet resulting in a 93.33% success. The highest accuracy was achieved when comparing kitchen sink and toilet flushing where only one kitchen sink event was misclassified as toilet flushing resulting in a 99.17% success.

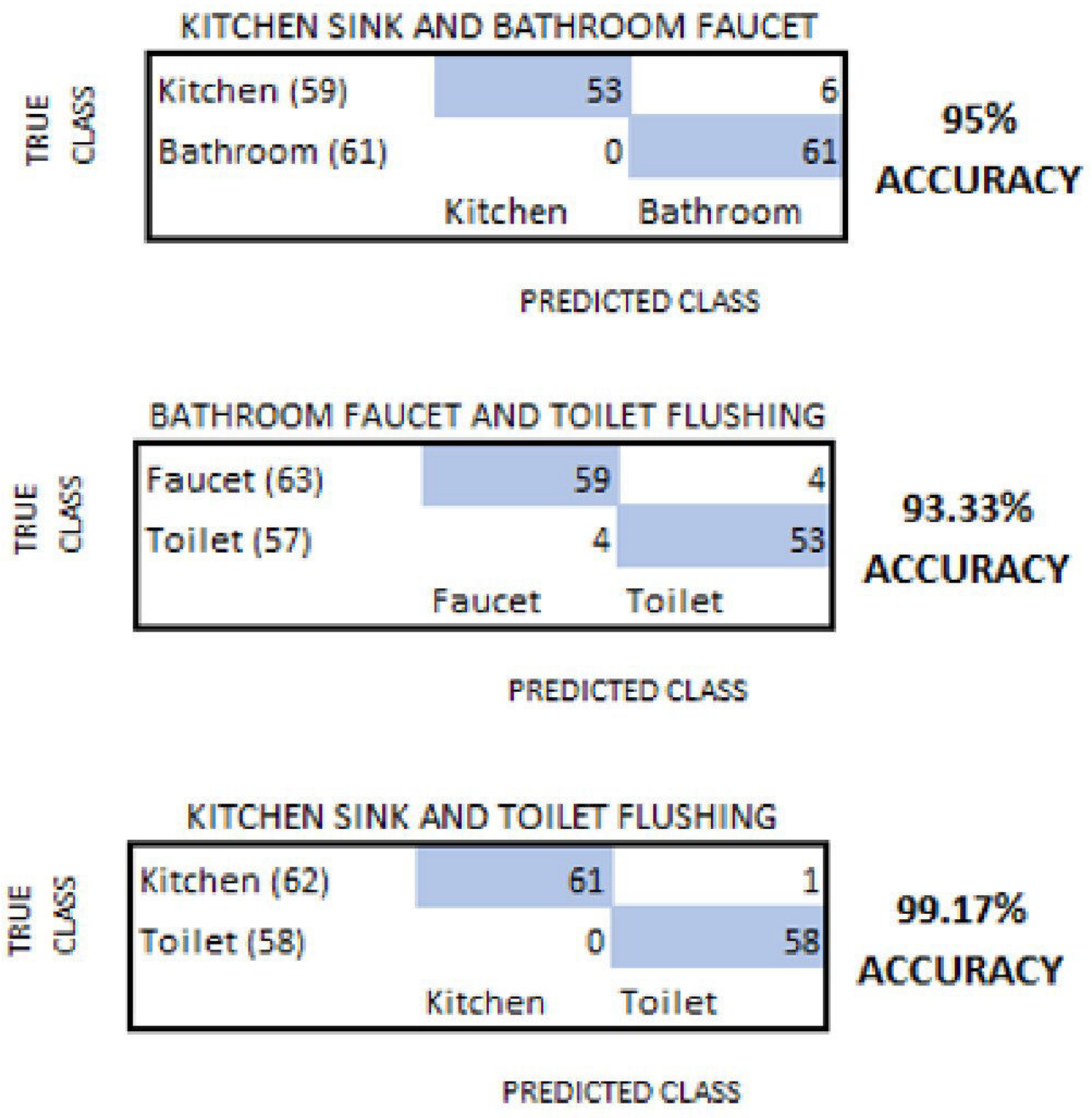

Fig. 10. Accuracy of Confusion Matrix

## A. *Comparative Study Using ML Classifiers*

We understand and acknowledged that the quantity of data was a potentially limiting factor that may have influenced the results, however, the fact should be considered that the second batch of data was observed and collected in a very controlled and monitored environment which may help the accuracy that was another limiting factor that shall need further testing. As a result, we evaluate the proposed model performance by comparing it with the performance of Logistic Regression (LR), Gaussian Naive Bayes (NB), Support Vector Machine (SVM), Random Forest (RF), Decision Tree (DT) and Neural Network (NN) classifier. The dataset was randomly split into training and test data while maintaining the class ratio among the class samples. Trained data was used to train each of the models we experimented with while test data was used for evaluating the performance of the models. To verify the consistency of the model, we experimented with each of the models with 5-fold cross-validation as the dataset is small containing only 90 samples in total, ensuring that each of the samples in the dataset would be considered both in training and test data. Therefore, in each fold: 80% of the data was used for training and the remaining 20% samples were used for evaluating model performance. We applied one versus rest method for all the classifiers since this is a multiclass classification problem and consider average performance across all three classes to measure accuracy, recall and precision. The algorithms were implemented using Python sci-kit-learn library with available hyperparameter options.

The neural network-based architecture consists of 4 layers including one input layer, two hidden layers and one output layer. We used the "ReLu" activation function in the hidden layers and the "sigmoid" function in the output layer, as this is a binary classification problem. "Adam" and "binary cross-entropy" were used for optimizer and loss function respectively. We implemented an early stopping method to stop training once the model performance stops improving on the test data. We selected validation loss to be monitored for early stopping and set minimum delta to 1e-3 (checks minimum change in the monitored quantity to qualify as an improvement) and patience to 5 (checks number of epochs that produced the monitored quantity with no improvement after which training will be stopped). The initial learning rate was set to 0.01.

The experiments are carried out on the Google CoLab platform with python 3.7 version. We implemented the experiment regarding neural networks on the Keras framework.

### *B. Comparative Findings*

We applied DT, RF, NB, LR, and NN classifiers to classify among hygiene classes: Kitchen Sink, Bathroom Faucet and Toilet Flushing. Table III demonstrates the results of the models in terms of accuracy, recall and precision. Both tree-based classifiers (DT and RF) outperforms other models by achieving highest accuracy (1.0) with a zero-standard deviation, recall score and and precision score respectively (1.0 $\pm$ 0.00) and (1.0 $\pm$ 0.00).

Other classifiers (LR, NB, and SVM) provide reasonable performance in terms of the evaluation metrics. However, NN shows poor performance (accuracy: 53%, recall 53%, and precision 34%) compared to other classifiers since training NN-based architecture requires large-scale samples to learning underlying patterns of the data. Finally, we can select either RF or DT for hygiene classification

TABLE III
EXPERIMENTAL RESULTS ANALYSIS OF DIFFERENT CLASSIFIERS

| Classifiers | Accuracy | Recall | Precision |
|---|---|---|---|
| DT | 1.0 $\pm$ 0.00 | 1.0 $\pm$ 0.00 | 1.0 $\pm$ 0.00 |
| RF | 1.0$\pm$0.00 | 1.0$\pm$0.00 | 1.0$\pm$0.00 |
| NB | 0.96$\pm$0.04 | 0.96$\pm$0.07 | 0.96$\pm$0.07 |
| LR | 0.89$\pm$0.05 | 0.89$\pm$0.13 | 0.89$\pm$0.11 |
| SVM | 0.91$\pm$0.04 | 0.91$\pm$0.12 | 0.92$\pm$0.10 |
| NN | 0.53$\pm$0.16 | 0.53$\pm$0.50 | 0.34$\pm$0.38 |

## V. DISCUSSION AND FUTURE DIRECTION

The results obtained through the experiments validate this research and the concept of using vibration sensors and machine learning techniques to monitor hygiene activities and automate the classification process. To overcome the limitation of the limited dataset, we conducted an experiment that was performed by comparing the general accuracy with the performance of Logistic Regression (LR), Gaussian Naive Bayes (NB), Support Vector Machine (SVM), Random Forest (RF), Decision Tree (DT) and Neural Network (NN) classifier provide high efficiency in terms of validation of both data and accuracy.

As far as challenges with network connectivity, the recommendation would be to try wired connection using Ethernet instead of solely relying on wireless for improving the connection and consistency. We also recommend verifying the collected data more often to identify any gaps whilst in the process of data collection to discard any data or recollect data. Collecting "real-world data" is critical for the success of the research, we need to consider minimizing external influences including people walking in or out of the designated areas, objects being placed or moved around the sensor. All of these external factors would create frequencies that would add some level of noise to the data, we recommend processing the dataset through a frequency analysis filter to remove any high/low frequencies that are not relevant to the hygiene activity itself.

Additionally, there are different shapes and sizes of faucets and toilets, including different builds and materials, we recommend considering testing for further validating the results. With the accuracy that we achieved in the controlled environment, we can confidently conclude that the proposed framework has the robustness in hygiene monitoring that justifies further research to advance. A foundation has been set through this research, we intend to implement advances further that shall potentially lead to a real-world implementation. In the future, we would like to intersect novel technologies including blockchain technology towards providing secure, transparent, non-corruptible collected data storing and sharing. We were inspired by our various successful studies including [30], [31] where we show the potential in storing medical health records or electronic health records (EHR) in the blockchain-based network. We also plan to extend our research on utilizing machine-learning techniques in the advanced level, Random Forest and Decision Tree classifiers were used in our recent work where we showed ML can be used to achieve the highest accuracy in Autism Spectrum Disorder Detection [32].

## VI. Conclusion

In this paper, we introduced a novel non-invasive framework for monitoring human hygiene using vibration sensors combination of a geophone sensor, a digitizer, and a cost-efficient computer board in a practical enclosure. Machine Learning techniques are adopted for classifying the collected datasets and measuring the accuracy in detecting hygiene activities. The proposed approach consistently achieved an accuracy of $\sim$95% for each scenario while Random Forrest and Decision Tree outperform other models by achieving the highest accuracy of (100%). The considerable accuracy indicates the research potential of the proposed method in both the healthcare and technology sectors. With the help of continued research and through a real-world application of the proposed solution, healthcare professionals will be able to keep track of their patients' daily hygiene behavior and detect any abnormalities that may be a cause for concern. Keeping the end goal of "helping our healthcare professionals be proactive rather than reactive" where the proposed framework would equip healthcare professionals with tools to address any growing health concerns before they become life-threatening for the patient and the community.

## References

[1] W. H. Organization. (2021) Ageing and health. world health organization. [Online]. Available: https://www.who.int/news-room/fact-sheets/detail/ageing-and-health

[2] U. C. Bureau. (2016) U.s. population aging slower than other countries, census bureau reports. [Online]. Available: https://www.census.gov/newsroom/press-releases/2016/cb16-54.html

[3] S. K. Mark MatherPaola. (2019) Fact sheet: Aging in the united states. [Online]. Available: https://www.prb.org/resources/fact-sheet-aging-in-the-united-states/

[4] M. K. A. Roosa Tikkanen. (2020) U.s. health care from a global perspective, 2019: Higher spending, worse outcomes? [Online]. Available: https://www.commonwealthfund.org/publications/issue-briefs/2020/jan/us-health-care-global-perspective-2019

[5] M. Auley, A. Martinez Guimera, D. Hodgson, N. Mcdonald, K. Mooney, A. Morgan, and C. Proctor, "Modelling the molecular mechanisms of aging," *Bioscience Reports*, vol. 37, p. BSR20160177, 01 2017.

[6] L. D. Harris-Kojetin, M. Sengupta, J. P. Lendon, V. Rome, R. Valverde, and C. Caffrey. (2019) Long-term care providers and services users in the united states, 2015-2016.

[7] N. I. on Aging. (2017) What is long-term care? [Online]. Available: https://www.nia.nih.gov/health/what-long-term-care

[8] C. for Disease Control and Prevention. (2016) National center for health statistics: Nursing home care. [Online]. Available: https://www.cdc.gov/nchs/fastats/nursing-home-care.htm

[9] E. Hygiene. (2020) Hygiene fast facts information on water-related hygiene. [Online]. Available: www.cdc.gov/healthywater/hygiene/fast-facts.html

[10] M. J. Hossain Faruk, H. Shahriar, M. Valero, F. L. Barsha, S. Sobhan, M. A. Khan, M. Whitman, A. Cuzzocrea, D. Lo, A. Rahman, and F. Wu, "Malware detection and prevention using artificial intelligence techniques," in *2021 IEEE International Conference on Big Data (Big Data)*, 2021, pp. 5369–5377.

[11] M. J. Hossain Faruk and M. Adnan, "Smart children management using data analytics, machine learning and iot," in *International Conference on Artificial Intelligence for Smart Community*, 12 2020.

[12] P. Gogas and T. Papadimitriou, "Machine learning in economics and finance," *Computational Economics*, vol. 57, 02 2021.

[13] M. J. Hossain Faruk, M. Valero, and H. Shahriar, "An investigation on non-invasive brain-computer interfaces: Emotiv epoc+ neuroheadset and its effectiveness," in *2021 IEEE 45th Annual Computers, Software, and Applications Conference (COMPSAC)*, 2021, pp. 580–589.

[14] M. Masum, H. Shahriar, H. Haddad, M. J. Hossain Faruk, M. Valero, M. Khan, M. Rahman, M. Adnan, A. Cuzzocrea, and F. Wu, "Bayesian hyperparameter optimization for deep neural network-based network intrusion detection," 12 2021.

[15] M. Masum, M. J. Hossain Faruk, H. Shahriar, K. Qian, D. Lo, and M. Adnan, "Ransomware classification and detection with machine learning algorithms," 01 2022.

[16] Y. Zhang, J. Wullems, I. D'Haeseleer, V. V. Abeele, and B. Vanrumste, "Bathroom activity monitoring for older adults via wearable device," in *2020 IEEE International Conference on Healthcare Informatics (ICHI)*, 2020, pp. 1–10.

[17] J. Chen, J. Zhang, A. Kam, and L. Shue, "An automatic acoustic bathroom monitoring system," in *2005 IEEE International Symposium on Circuits and Systems (ISCAS)*, 2005, pp. 1750–1753 Vol. 2.

[18] Y. Hao, G. Zhang, J. Jiang, F. Shangguan, and Z. Zhang, "Design and implementation of an intelligent kitchen safety monitor," in *2020 IEEE 9th Joint International Information Technology and Artificial Intelligence Conference (ITAIC)*, vol. 9, 2020, pp. 93–97.

[19] K. Jimi, H. Seto, and A. Kajiwara, "Bathroom monitoring with fast-chirp modulation millimeter-wave uwb radar," in *2020 IEEE Radio and Wireless Symposium (RWS)*, 2020, pp. 134–137.

[20] J. Fagert, M. Mirshekari, S. Pan, P. Zhang, and H. Y. Noh, "Monitoring hand-washing practices using structural vibrations," *Structural Health Monitoring-an International Journal*, 2017.

[21] A. C. J. Klein Brinke and P. J. M. Havinga, "Personal hygiene monitoring under the shower using wifi channel state information," *Computer Human Interaction IoT Appllication*, 2021.

[22] M. Hotchkiss. (2022) What is raspberry shake? — iris. [Online]. Available: https://www.iris.edu/

[23] A. Barzilai, "Improving a geophone to produce an affordable, broadband seismometer," *Mechanical Engineering, Stanford University January*, vol. 25, 2000.

[24] P. Scanlon, R. Henriksen, and J. Allen, "Approaches to electromagnetic induction," *American Journal of Physics*, vol. 37, no. 7, pp. 698–708, 1969.

[25] Grafana Labs, "Grafana," 2018. [Online]. Available: https://grafana.com/

[26] Influxdata Inc, "InfluxDB," 2019. [Online]. Available: https://www.influxdata.com/

[27] S. S. S. B. L. Edemekong PF, Bomgaars DL, "Activities of daily living." Treasure Island (FL).

[28] A. Fabre, O. Saaid, F. Wiest, and C. Boucheron, "Current controlled bandpass filter based on translinear conveyors," *Electronics Letters*, vol. 31, no. 20, pp. 1727–1728, 1995.

[29] C. Chen, L. Song, C. Bo, and W. Shuo, "A support vector machine with particle swarm optimization grey wolf optimizer for network intrusion detection," in *2021 International Conference on Big Data Analysis and Computer Science (BDACS)*, 2021, pp. 199–204.

[30] M. J. Hossain Faruk, H. Shahriar, M. Valero, S. Sneha, S. Ahamed, and M. Rahman, "Towards blockchain-based secure data management for remote patient monitoring," *IEEE International Conference on Digital Health (ICDH)*, 2021.

[31] M. J. Hossain Faruk, "Ehr data management: Hyperledger fabric-based health data storing and sharing," *The Fall 2021 Symposium of Student Scholars*, 2021. [Online]. Available: https://www.researchgate.net/publication/356507747

[32] M. Masum, I. Nur, M. J. Hossain Faruk, M. Adnan, and H. Shahriar, "A comparative study of machine learning-based autism spectrum disorder detection with feature importance analysis," in *COMPSAC 2022: Computer Software and Applications Conference*, 03 2022. [Online]. Available: https://www.researchgate.net/publication/359081817